\documentclass[10pt,twocolumn,letterpaper]{article}

\usepackage[datasets]{wacv}      % To produce the REVIEW version for the algorithms track
\usepackage{mathrsfs}
\usepackage{tabularx}
\usepackage{booktabs}
\usepackage{multirow}
\usepackage[]{mdframed}
\usepackage{fvextra}

\definecolor{correctgreen}{RGB}{0,120,70}
\definecolor{wrongred}{RGB}{180,40,40}
\definecolor{neutralblue}{RGB}{0,100,180}

\definecolor{wacvblue}{rgb}{0.21,0.49,0.74}
\usepackage[pagebackref,breaklinks,colorlinks,allcolors=wacvblue]{hyperref}

\def\wacvPaperID{*****} % *** Enter the WACV Paper ID here
\def\confName{WACV}
\def\confYear{2027}

\title{Adversarial Vulnerability of Relational Reasoning in LVLMs}
\author{Kawshik Banerjee and Khaled Mohammed Saifuddin\\
Southern Illinois University, Carbondale\\
{\tt\small \{kawshik.banerjee, k.saifuddin\}@siu.edu}
}

\begin{document}

\newcommand\blfootnote[1]{
  \begingroup
  \renewcommand\thefootnote{}\footnote{#1}
  \addtocounter{footnote}{-1}
  \endgroup
}
\maketitle
\blfootnote{Preprint}

\begin{abstract}
Large Vision-Language Models (LVLMs) are increasingly deployed in safety-critical applications, like medical image understanding, yet remain vulnerable to adversarial image perturbations. While prior work has primarily focused on LVLMs' overall robustness, we investigate whether adversarial attacks disproportionately affect relational understanding compared to object recognition in Visual Question Answering (VQA) tasks. Evaluating six LVLMs, including adversarially fine-tuned defense models, across four benchmarks requiring object-level and relational visual reasoning under adversarial attacks, we frequently observe greater accuracy loss in relation-based tasks than in object-based ones. We quantify this disparity by presenting two complementary diagnostic metrics defined as the ratio and gap of these two capability-specific drops under attack. A controlled image-paired evaluation provides further evidence of this disparity while reducing potential confounding from differences in the evaluated images, and a comparison against Gaussian noise reaffirms that the vulnerability is specific to adversarial optimization rather than general image corruption. A similar relational gap in domain-specific medical LVLMs highlights potential concern that capability-level disparities warrant further investigation in medical image understanding. Finally, our analysis of three adversarially fine-tuned models reveals a persistent residual capability-level gap, motivating more fine-grained robustness evaluation strategies for LVLMs. 
% Codes are available at: \href{https://anonymous.4open.science/r/Relational-Blind-Spot-3125/}{anonymous.4open.science/r/Relational-Blind-Spot-3125}
\end{abstract}

\section{Introduction}
\label{sec:intro}
\begin{figure*}[htb]
\centering
\includegraphics[height=0.55\textwidth, width=0.9\textwidth]{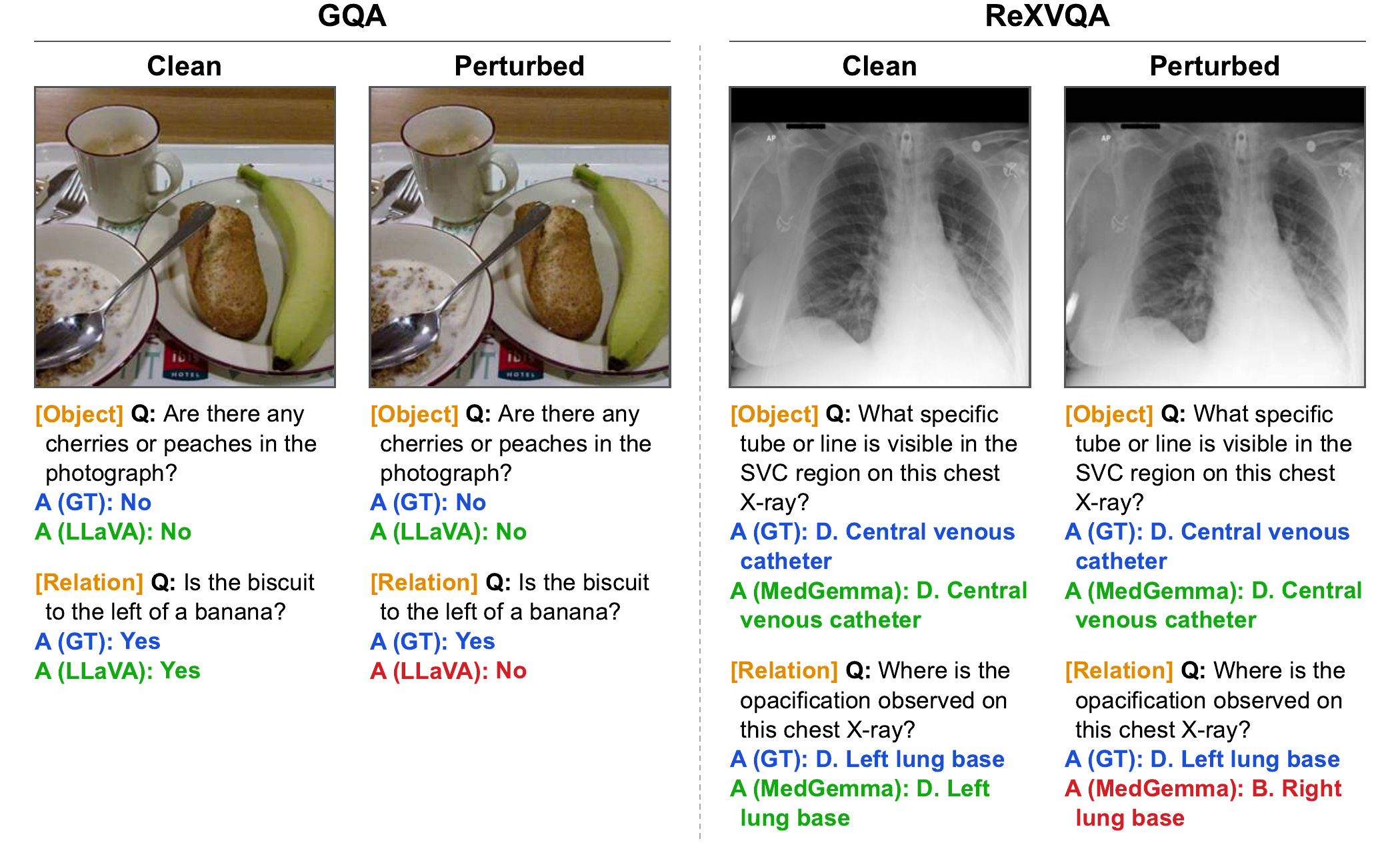}
\caption{Adversarial perturbations disproportionately degrade relational understanding. LLaVA-1.5-7B \cite{liu2024improved} answers questions from the GQA \cite{hudson2019gqa} (left panel) and MedGemma-1-27B \cite{sellergren2025medgemma} answers from ReXVQA \cite{pal2025rexvqa} (right panel) with clean (left of each panel) and PGD-perturbed \cite{madry2017towards} (right of each panel) images as input. Under attack, object recognition remains correct while relational understanding fails. The ground-truth answers are highlighted in \textcolor{neutralblue}{blue}; correct and incorrect responses from the models are highlighted in \textcolor{correctgreen}{green} and \textcolor{wrongred}{red}, respectively.}
\label{fig:sample}
\vspace{-5mm}
\end{figure*}
% Paragraph 1: LVLMs deployed in safety-critical settings, adversarial vulnerability known
Large vision-language models (LVLMs) have made significant progress in recent times in various vision-grounded tasks like image captioning \cite{yang2023exploring}, visual question answering (VQA) \cite{alayrac2022flamingo}, visual-grounded reasoning \cite{chen2024spatialvlm}, text-to-image generation \cite{pmlr-v139-ramesh21a}, along with safety-critical settings like autonomous driving \cite{tian2024drivevlm}, and clinical diagnosis \cite{li2023llava, anatomyvlm}. Consequently, LVLMs have rapidly transitioned from research prototypes to widely deployed AI systems \cite{gpt4vision, gemini_announcement, raza2025industrial}, with multimodal capabilities integrated into consumer applications \cite{blv, gpt4vision}, enterprise platforms \cite{you2024ferret, brynjolfsson2025generative}, healthcare \cite{garcia2024artificial, li2023llava}, finance \cite{lopez2026can}, and other real-world settings. 
% Such rapid adoption has outpaced the field's understanding of the systems' vulnerabilities \cite{schaeffer2025failures}. Prior studies have shown how these vulnerabilities are exploited and how we can defend our systems against them \cite{ye2025survey}. 
Such rapid adoption has outpaced the field's understanding of their vulnerabilities, including susceptibility to adversarial inputs \cite{zhao2023evaluating, qi2024visual}, hallucination \cite{li2023evaluating}, and distributional shift \cite{cai2024benchlmm}. Adversarial image perturbations--imperceptible modifications to inputs that cause drastic changes in model outputs--represent a particularly concerning threat to deployed LVLMs \cite{szegedy2013intriguing, zhao2023evaluating}. Prior studies have shown how such adversarial attacks are exploited \cite{ye2025survey, liu2025survey} and how we can defend our systems against them \cite{tecoa, schlarmann2024robust, slade}. 

% Paragraph 2 — Gap: prior work measures overall robustness, no capability-level breakdown
However, current literature takes a holistic approach while assessing the degree and impact of adversarial attacks through aggregate metrics like Consensus-based Image Description Evaluation (CIDEr) score or VQA accuracy as a whole \cite{ye2025survey, liu2025survey}. A model that retains object recognition while losing all spatial understanding and a model that degrades uniformly under such attacks can perform nearly identically on these metrics on standard benchmarks, yet represent fundamentally different failure modes \cite{xu2021robust, benz2021robustness}. 
% as a whole while containing very different forms of vulnerabilities.
% The failure mode is as crucial as the failure rate — an attack can amplify an LVLM's existing relational weaknesses to generate seemingly plausible yet incorrect outputs (e.g., confusing left and right lungs while diagnosing a radiograph \cite{kathait2024assessing, meddeb2025evaluating}), compounding risks that are already present in unperturbed settings, a pattern we confirm empirically across multiple benchmarks and models in this work For an adversarial attack that amplifies an existing weak capability, holistic evaluations will consistently understate the risk.   
The failure mode here is as crucial as the failure rate. An attack can amplify existing relational weaknesses in LVLMs, producing seemingly plausible yet incorrect outputs (e.g., confusing left and right lungs while diagnosing a radiograph \cite{kathait2024assessing, meddeb2025evaluating}). We confirm this amplification empirically across multiple benchmarks and models in this work. For adversarial attacks that disproportionately affect an existing weak capability, holistic evaluations will understate the risk to some extent.

% Paragraph 3 — Our finding: relational understanding disproportionately vulnerable, rho metric, controlled experiment

This work identifies a specific and previously underexplored failure mode of this risk: adversarial attacks disproportionately degrade relational understanding compared to object recognition (see Fig. \ref{fig:sample} for a sample example). We evaluate two general-purpose LVLMs, one domain-specific fine-tuned model, and three adversarially fine-tuned defense models on VQA benchmarks involving object- and relation-level visual reasoning questions under $L_\infty$ and $L_2$ Projected Gradient Descent (PGD) \cite{madry2017towards} attacks. Our results show a tendency toward greater degradation in relational accuracy compared to object recognition, although the pattern varies across models and benchmarks. To quantify this disparity, we present two diagnostic metrics, $\rho$ and $G$, defined as the ratio and gap of relation- to object-level understanding degradation under attacks. Multiple controlled experiments further indicate that this gap is statistically significant and induced by adversarial perturbations. We also evaluate a domain-specific medical LVLM on clinical benchmarks and observe a comparable relational gap, suggesting that this vulnerability may extend to safety-critical settings, where such failures could have more serious consequences.
% Paragraph 4 — Existing defenses fail to close relational gap
Furthermore, the defense models that have been fine-tuned against adversarial attacks also fail to address this relational gap. Evaluating multiple such defense models, we observe that while each improves overall performance compared to the vanilla model, the ratios $\rho$ and gap $G$ remain high under attack. This constitutes a critical blind spot while optimizing defense models against perturbations. 

Our contributions are as follows:
\begin{itemize}
    \item We identify a previously underexplored vulnerability in LVLMs: adversarial attacks disproportionately degrade relational understanding relative to object recognition, a disparity that persists across models and benchmarks.
     \item We evaluate three LVLMs and three adversarially fine-tuned defense models across four VQA benchmarks under both $L_\infty$ and $L_2$ PGD attacks, finding that existing defenses fail to close the relational gap across attack norms, exposing a critical blind spot in current robustness strategies.
    \item We present \textit{Capability Disparity Ratio}, $\rho$, and \textit{Capability Disparity Gap}, $G$, two complementary diagnostic metrics that quantify capability-specific adversarial vulnerability as the ratio and gap of relational to object-level accuracy drops under attack, enabling a systematic comparison across models and attack norms.
    \item We conduct a controlled image-paired evaluation that rules out image-level difficulty as a confound, supporting that the relational gap is statistically significant under adversarial attack.
\end{itemize}

\section{Related Work}
\subsection{Large Vision-Language Models} 
Modern LVLMs primarily follow a modular design comprising a vision encoder and a large language model (LLM) backbone, connected via a projection layer \cite{liu2023visual, pmlr-v235-karamcheti24a}, although alternative architectures \cite{alayrac2022flamingo} and encoder-free variants \cite{team2026gemma} also exist. A representative example is LLaVA-1.5 \cite{liu2024improved}, which couples a CLIP ViT-L/14 vision encoder \cite{radford2021learning} with a linear projector and a Vicuna backbone \cite{vicuna2023}, an architecture we adopt as a primary baseline in this work. However, many LVLM instruction-tuning pipelines freeze the vision encoder while updating the projector and backbone, which potentially leaves the encoder lacking defenses against adversarial inputs \cite{ye2025survey}.

\subsection{Adversarial Robustness of Vision-Language Models}
% Cover adversarial attacks on VLMs and
% adversarial defenses (FARE, TeCoA, SLADE). End with gap: none of these
% distinguish relational vs object vulnerability under attack — all prior work
% measures overall robustness without capability-level breakdown

The integration of the visual inputs introduces new attack surfaces in LLMs \cite{lee2025does, ye2025survey, li2024images}. While text-based attacks search for distinct token spaces \cite{zou2023universal}, visual attacks can follow gradients directly with human-imperceptible perturbations \cite{sch2023adv, madry2017towards}. Prior work shows this is a vulnerability for both open-weight and proprietary models alike \cite{dong2023robust}, and a single perturbation can generalize and be transferable \cite{luo2024image, wang2024stop, gao2024adversarial}. Such vulnerabilities can be exploited to bypass models' safety alignments. LVLMs can thus be made to generate toxic results \cite{NEURIPS2023_c1f0b856, qi2024visual}, bypass specific instructions \cite{gong2025figstep}, bypass text-based safeguards \cite{shayegani2024jailbreak} or to attack the model itself \cite{qraitem2024vision}.

The practical implications of these exploits scale with deployment context. Models fine-tuned for medical domains, such as LLaVA-Med \cite{li2023llava}, Med-Flamingo \cite{moor2023med}, or the recent MedGemma \cite{sellergren2025medgemma}, are being positioned for diagnostic tasks, yet medical imaging is not safe from such vulnerabilities \cite{PromptSmooth, budathoki2025adversarial}. Defense strategies against such perturbations can be broadly categorized as inference- and training-time methods \cite{ye2025survey}. Inference-time methods leave the model weights as-is \cite{pmlr-v162-nie22a, wang2024adashield, xu2024cross, gou2024eyes}.
% Training-time defenses instead safeguard the encoder. TeCoA \cite{tecoa} performs supervised fine-tuning of CLIP with cross-modal contrastive loss and improves robustness. FARE \cite{schlarmann2024robust} eliminates over-reliance on text by minimizing the squared $L_2$ distance between the fine-tuned encoder's embedding of the perturbed image and the original CLIP encoder's embedding of the clean image. However, a single such objective across the whole image can be insufficient for retaining local details under stronger perturbation \cite{slade}. SLADE \cite{slade} addresses this with a dual symmetric contrastive objective applied simultaneously at global and local (patch) levels. However, all of such defense models evaluate performance as a whole, and none isolates which kind of information is more susceptible to these perturbations.
Training-time defenses safeguard the vision encoder directly. TeCoA \cite{tecoa} fine-tunes CLIP with a cross-modal contrastive loss to improve zero-shot robustness. FARE \cite{schlarmann2024robust} fine-tunes the vision encoder so that its embedding of a perturbed image stays close to that of the original frozen one of the clean image. SLADE \cite{slade} extends this with a dual symmetric contrastive objective at both global and local patch levels. However, all such defenses evaluate performance holistically, and none examines whether relational reasoning is disproportionately affected.

\subsection{Relational and Spatial Reasoning in LVLMs}
% Cover relational and spatial reasoning benchmarks (GQA, Spatial-MM, DORI)
% and how LVLMs process inter-object relationships under clean conditions.
% End with gap: prior work studies relational reasoning only on clean images —
% no work examines whether adversarial perturbation specifically exploits
% relational understanding as a distinct vulnerability.
Prior studies assessing LVLMs on clean images highlighted that relational and spatial understanding is a specific and inherent limitation \cite{khangaonkar2026multimodal, spatialqa}, and current assessments often treat spatial or orientation understanding as a secondary evaluation target \cite{tasnim2025seeing}. LVLMs often fail at image-text matching when captions contain identical words in different compositional settings \cite{thrush2022winoground} and perform poorly in orientation tasks \cite{vsr}. However, despite remarkable progress in LVLMs, this gap persists \cite{chen2024spatialvlm, fu2024blink, wang2024picture}. Crucially, all prior assessments of relational and spatial reasoning are conducted exclusively on clean, unperturbed images. Whether adversarial perturbations selectively exploit relational understanding as a distinct vulnerability above and beyond their general degradation of model performance remains entirely unexplored. We address this gap directly.

\section{Methodology}
\label{sec:methodology}

\subsection{Problem Formulation}
\label{sec:problem}
Existing adversarial robustness evaluations of LVLMs on VQA benchmarks typically report aggregate accuracy over all question types (including object- and relation-oriented questions), blending object-centric and relational reasoning capabilities into a single score. This obscures whether adversarial perturbations affect these two capabilities equally or disproportionately. To enable a fine-grained analysis, we partition the evaluation set by capability type and define separate accuracy measures for object-centric and relational questions.

Let $f_{\theta}: \mathcal{X} \times \mathcal{Q} \rightarrow \mathcal{Y}$ denote an LVLM parameterized by $\theta$, where $\mathcal{X}$, $\mathcal{Q}$, and $\mathcal{Y}$ represent the image, question, and answer spaces, respectively, and $\mathcal{M}$ is the set of LVLMs we evaluate. Given an image--question pair $(x_i, q_i)$, the model predicts $\hat{y}_i = f_{\theta}(x_i, q_i)$. We consider an evaluation set $\mathcal{E} = \{(x_i, q_i, y_i, c_i)\}_{i=1}^{N}$, where $y_i$ is the ground-truth answer and $c_i \in \{\mathrm{obj}, \mathrm{rel}\}$ denotes the capability required to answer the question. We accordingly partition $\mathcal{E}$ into $\mathcal{E}_{\mathrm{obj}} = \{(x_i, q_i, y_i, c_i) \in \mathcal{E} : c_i = \mathrm{obj}\}$ and $\mathcal{E}_{\mathrm{rel}} = \{(x_i, q_i, y_i, c_i) \in \mathcal{E} : c_i = \mathrm{rel}\}$. Object-centric questions in $\mathcal{E}_{\mathrm{obj}}$ require identifying individual entities or their attributes (e.g., ``Are there any cherries in the plate?''), while relational questions in $\mathcal{E}_{\mathrm{rel}}$ require reasoning over spatial configurations involving the entities (e.g., ``Is the biscuit to the left of a banana?''). For a subset $\mathcal{S} \subseteq \mathcal{E}$, the clean accuracy is: 
\begin{equation} 
\operatorname{Acc}_{\mathrm{clean}}(\mathcal{S}) = \frac{1}{|\mathcal{S}|} \sum_{(x_i,q_i,y_i,c_i) \in \mathcal{S}} \mathbb{1}\left[f_{\theta}(x_i, q_i) = y_i\right] 
\end{equation} 

The capability-specific accuracies are $\operatorname{ObjAcc}_{\mathrm{clean}} = \operatorname{Acc}_{\mathrm{clean}}(\mathcal{E}_{\mathrm{obj}})$ and $\operatorname{RelAcc}_{\mathrm{clean}} = \operatorname{Acc}_{\mathrm{clean}}(\mathcal{E}_{\mathrm{rel}})$. Overall adversarial accuracy does not reveal whether these two capabilities degrade at the same rate. We therefore investigate whether adversarial image perturbations produce a systematically larger loss in relational accuracy than in object-centric accuracy. To ensure this comparison is not confounded by differences in the images associated with the two question types, we describe both the perturbation protocol and a controlled evaluation procedure in the following subsections.

\subsection{Adversarial and Noise Perturbations}
\label{sec:perturbation}

Measuring capability-specific vulnerability requires perturbing inputs in a controlled and principled manner. We consider two types of image perturbations: adversarial perturbations optimized to maximize model error, and random Gaussian noise as a non-adversarial control. The contrast between these two allows us to isolate whether any observed relational gap is driven by adversarial optimization specifically, or whether it reflects a general sensitivity to image corruption. Our primary evaluation uses $L_\infty$-bounded adversarial perturbations: 
\begin{equation} 
x' = x + \delta, \quad \|\delta\|_\infty \leq \epsilon, 
\end{equation} 
where $\epsilon = 4/255$ following standard evaluation practice. The adversarial image $x'$ is obtained via PGD \cite{madry2017towards}:
% \begin{equation} 
% \delta^{t+1} = \Pi_\epsilon \left( \delta^t + \alpha \cdot \text{sign}\left(\nabla_\delta \mathcal{L}\bigl(f_{\theta}(x + \delta^t, q), y\bigr)\right) \right), 
% \end{equation} 
\begin{equation}
\delta^{t+1} = \Pi_\epsilon \left( \delta^t + \alpha \cdot \text{sign}\left(\nabla_\delta \mathcal{L}_{\text{feat}}(\delta^t)\right) \right),
\end{equation}
% with $x' = x + \delta^T$ after $T$ iterations, where $\Pi_\epsilon$ denotes projection onto the $L_\infty$ ball of radius $\epsilon$, $\alpha$ is the step size, and $\mathcal{L}$ is the task loss. We maximize this task loss with respected to the ground truth answer.
with $x' = x + \delta^T$ after $T$ iterations, where $\Pi_\epsilon$ denotes projection onto the $L_\infty$ ball of radius $\epsilon$, $\alpha$ is the step size, and
\begin{equation}
    \mathcal{L}_{\text{feat}}(\delta) = -\frac{1}{P} \sum_{p=1}^{P} \mathrm{cos} (g_\phi(x + \delta)_p, g_\phi(x)_p)
\end{equation}
is the negative cosine similarity between clean and perturbed patch-token features over $P$ tokens of the vision encoder $g_\phi$. Since this objective depends only on the image, the perturbation is identical across all questions associated with the image, making it capability-agnostic. A larger relational degradation under PGD would not, by itself, establish an adversarially induced vulnerability: relational questions might also be more sensitive to arbitrary pixel-level corruption. We therefore use Gaussian noise as a non-adversarial control: 
\begin{equation} 
\delta_{\text{Gaussian}} \sim \mathcal{N}(0, \sigma^2), \quad \sigma = \frac{\epsilon}{2}, \quad \text{clipped to } [-\epsilon, \epsilon] 
\end{equation} 

Unlike PGD, Gaussian noise introduces random pixel-level distortions with no optimization objective. Comparing PGD with this control separates attack-directed degradation from sensitivity to generic input corruption, since evidence of relational vulnerability requires the relation-object disparity to be substantially stronger under PGD than under Gaussian noise.

\subsection{Quantifying Relational Vulnerability}
\label{sec:metric}

Having defined the evaluation sets and perturbation protocol, we now introduce metrics to quantify how much adversarial attacks disproportionately affect relational reasoning. A simple comparison of accuracy drops would conflate the difficulty of each capability type with its adversarial vulnerability. We therefore first measure the relative drop from each capability's own clean baseline. Let $\operatorname{ObjAcc}_{\mathrm{adv}}$ and $\operatorname{RelAcc}_{\mathrm{adv}}$ denote accuracies on adversarially perturbed images for object- and relation-based questions, respectively, with the corresponding drops defined as:
\begin{equation} 
\label{eqn:del_obj_acc} 
\Delta\operatorname{ObjAcc} = \operatorname{ObjAcc}_{\mathrm{clean}} - \operatorname{ObjAcc}_{\mathrm{adv}} 
\end{equation} 
\begin{equation} 
\label{eqn:del_rel_acc} 
\Delta\operatorname{RelAcc} = \operatorname{RelAcc}_{\mathrm{clean}} - \operatorname{RelAcc}_{\mathrm{adv}} 
\end{equation} 

We introduce $\rho$, the \textit{Capability Disparity Ratio}, as a diagnostic metric: 
\begin{equation} 
\label{eqn:ratio} \rho = \Delta\operatorname{RelAcc} / \Delta\operatorname{ObjAcc}
\end{equation} 

The ratio has a direct interpretation: $\rho = 1$ indicates equal accuracy degradation, $\rho > 1$ indicates greater relational degradation, and $\rho < 1$ indicates greater object-level degradation. Because $\rho$ can become unstable when $\Delta\operatorname{ObjAcc}$ is close to zero, we additionally report a complementary \textit{Capability Disparity Gap}: 
\begin{equation} 
\label{eqn:absolute_disparity}
G = \Delta\operatorname{RelAcc} - \Delta\operatorname{ObjAcc} 
\end{equation} 

Thus, $\rho$ captures the proportional disparity, while $G$ provides a stable linear disparity when the denominator is small.

\subsection{Defense Evaluation}
\label{sec:defense}

Even if adversarial attacks disproportionately degrade relational reasoning, existing defenses may already address this gap. We therefore ask whether current adversarial defenses recover relational reasoning as effectively as object recognition. Because defense models may differ in their clean accuracies, raw defended accuracy is not directly comparable across methods. We therefore measure the fraction of each model's clean performance that remains unrecovered under attack. For each defense model $d \in \mathcal{M_\mathrm{def}}$, where $\mathcal{M_\mathrm{def}} \subset \mathcal{M}$, we define the normalized residual gap as: 
\begin{equation} 
\label{eqn:ng_obj} 
\operatorname{NG}_{\mathrm{obj}}^{d} = \frac{\operatorname{Acc}_{\mathrm{obj, clean}} - \operatorname{Acc}^{d}_{\mathrm{obj, adv}}}{\operatorname{Acc}_{\mathrm{obj, clean}}} \times 100 
\end{equation} 

\begin{equation} 
\label{eqn:ng_rel} 
\operatorname{NG}_{\mathrm{rel}}^{d} = \frac{\operatorname{Acc}_{\mathrm{rel, clean}} - \operatorname{Acc}^{d}_{\mathrm{rel, adv}}}{\operatorname{Acc}_{\mathrm{rel, clean}}} \times 100 
\end{equation} 

where $\operatorname{Acc}^{d}_{\mathrm{obj, adv}}$ and $\operatorname{Acc}^{d}_{\mathrm{rel, adv}}$ denote the object and relational accuracies of defense model $d$ under adversarial attack. Lower values indicate better preservation of clean performance. The residual relational disadvantage after applying defense $d$ is: 
\begin{equation} 
\label{eqn:normalized_gap}
G_{\mathrm{def}}^{d} = \operatorname{NG}_{\mathrm{rel}}^{d} - \operatorname{NG}_{\mathrm{obj}}^{d} \end{equation} 
A positive value of $G_{\mathrm{def}}^{d}$ indicates that the defense recovers a smaller proportion of relational performance than of object-level performance, suggesting that existing defenses inadequately protect relational reasoning even when they partially restore overall accuracy.

\subsection{Controlled Evaluation Protocol}
\label{sec:controlled}

A direct comparison between object and relational questions over the full evaluation set may be confounded by differences in the underlying images, as relational questions may disproportionately occur in visually crowded or compositionally complex scenes. To address this confound, we construct an image-paired subset containing only images associated with an equal non-zero number of object-centric and relational questions. Our assumption here is that a complex and visually cluttered scene with many entities should ideally have a higher number of questions on the entities and the relationships between them, and so an image associated with an equal number of object- and relation-based questions with an identical capability-agnostic attack should pose a similar difficulty level for the models. For an image $x$ and capability $c \in \{\mathrm{obj}, \mathrm{rel}\}$, let $\mathcal{Q}_{c}(x) = \{(q_i, y_i) : (x_i, q_i, y_i, c_i) \in \mathcal{E},\; x_i = x,\; c_i = c\}$. The paired image set is then: 
\begin{equation}
\mathcal{X}_{\mathrm{paired}}
=
\left\{
x :
\left|\mathcal{Q}_{\mathrm{obj}}(x)\right|
=
\left|\mathcal{Q}_{\mathrm{rel}}(x)\right|
> 0
\right\}.
\end{equation} 
For every image in $\mathcal{X}_{\mathrm{paired}}$, both capability types are evaluated on clean images and their identical adversarially perturbed counterparts, controlling for image-level visual complexity. We report results on both the complete evaluation set and this subset. A consistent relational degradation across both provides stronger evidence that the observed disparity reflects a genuine capability-level vulnerability.

\subsection{Statistical Validation}
\label{sec:bootstrap}

% Observed differences in aggregate accuracy may depend on the sampled questions. We therefore estimate uncertainty in both $\rho$ and $G$ using nonparametric bootstrap resampling. To preserve dependencies among questions associated with the same image, the bootstrap is performed at the image level: images are sampled with replacement, and all eligible questions associated with each sampled image are retained. For each of $B = 1000$ bootstrap iterations, we compute $\hat{\rho}^{(b)}$ and $\hat{G}^{(b)}$ from the resampled evaluation set using pre-computed per-question accuracy scores — without rerunning the model. The 95\% confidence intervals are: 
% \begin{equation} 
% \text{CI}_{95\%}(\rho) = \left[\hat{\rho}_{2.5}, \hat{\rho}_{97.5}\right], \qquad \text{CI}_{95\%}(G) = \left[\hat{G}_{2.5}, \hat{G}_{97.5}\right] 
% \end{equation} 

% where the subscripts denote the 2.5th and 97.5th percentiles of the respective bootstrap distributions. We regard the evidence as statistically reliable when the 95\% confidence interval for $\rho$ lies above one and the corresponding interval for $G$ lies above zero. Reporting both intervals prevents conclusions from being driven by an unstable ratio denominator.

Having established the image-paired evaluation protocol, we now assess the statistical reliability of the observed relational gap on this controlled subset. Observed differences in aggregate accuracy may depend on which images are included in the evaluation. We therefore estimate uncertainty in $G$ using nonparametric bootstrap resampling. To preserve dependencies among questions associated with the same image, the bootstrap is performed at the image level: images are sampled with replacement, and all questions retained by the subset construction of section \ref{sec:controlled} are kept for each sampled image. Because each image contributes equal numbers of object- and relation-based questions, every resample preserves the 1:1 type ratio exactly, and thus no bootstrap draw can shift the object-relation composition of the evaluation set. For each of $B = 10{,}000$ bootstrap iterations $b = 1, \ldots, B$, we compute $\hat{G}^{(b)}$ from the resampled evaluation set using pre-computed per-question accuracy scores,  without rerunning the model. The 95\% confidence interval is: $ \text{CI}_{95\%}(G) = \left[\hat{G}_{2.5}, \hat{G}_{97.5}\right] $, where the subscripts denote the 2.5th and 97.5th percentiles of the bootstrap distribution. We regard the evidence as statistically reliable when the 95\% confidence interval for $G$ lies above zero. 

We additionally perform a paired permutation test on $G$, using $B=10{,}000$ permutations generated by randomly exchanging clean and adversarial labels within images. The two-sided $p$-value is computed as $p = \frac{k+1}{B+1}$, where $k$ is the number of permutations satisfying $|G^{(b)}| \geq |\hat{G}|$ for $b = 1 \dots B$.

\section{Experiments}
\subsection{Models}

We evaluate two general purpose LVLMs: LLaVA-1.5 (\texttt{llava-1.5-7b}) \cite{liu2024improved} and Gemma-4 (\texttt{gemma-4-26B-A4B-it}) \cite{team2026gemma}. For adversarially fine-tuned defense models, we evaluate SLADE \cite{slade}, TeCoA \cite{tecoa}, and FARE \cite{schlarmann2024robust}; all based on LLaVA. We use TeCoA and FARE as implemented in \cite{schlarmann2024robust} at $\epsilon=4/255$. We fine-tuned SLADE on the full ImageNet \cite{deng2009imagenet} at $\epsilon=4/255$ on $L_\infty$ following \cite{slade}. In the following sections, we report the performance of the version of SLADE that we fine-tuned. As the domain-specific model, we assess MedGemma-1 (\texttt{medgemma-27b-it}) \cite{sellergren2025medgemma}, a Gemma-based model fine-tuned for medical diagnosis.

\subsection{Datasets}

We evaluate across four VQA benchmarks spanning general visual reasoning, spatial understanding, and medical imaging, summarized in Table~\ref{tab:datasets}. GQA~\cite{hudson2019gqa} provides explicit object- and relation-level question splits (\texttt{val\_balanced}), making it our primary benchmark for capability-level analysis. Spatial-MM~\cite{shiri-etal-2024-empirical} focuses on spatial reasoning with single- and dual-object questions. MedXpertQA~\cite{MedXpertQA} and ReXVQA~\cite{pal2025rexvqa} extend evaluation to the medical domain, covering understanding vs.\ reasoning tasks and object detection vs.\ spatial reasoning tasks.

\begin{table}[htb]
    \centering
    \caption{Data statistics. Open: open-ended; MCQ: multiple-choice questions.}
    \label{tab:datasets}
    \setlength{\tabcolsep}{4pt}
    \renewcommand{\arraystretch}{1.05}
    \small
    \begin{tabular}{l l r r}
        \toprule
        \textbf{Dataset} & \textbf{Type} & 
        \textbf{\#Images ($\lvert \mathcal{X} \rvert$)} & \textbf{\#Questions ($\lvert \mathcal{Q} \rvert$)} \\
        \midrule
        GQA         & Open & 9,408  & 77,198 \\
        Spatial-MM  & MCQ  & 999    & 1,576 \\
        % DORI       & MCQ  & 16,802 & 33,240 \\
        MedXpertQA  & MCQ  & 2,858  & 2,000 \\
        ReXVQA      & MCQ  & 16,873 & 34,863 \\
        \bottomrule
    \end{tabular}
\end{table}
% {\small
% $^\dagger$ Open: open-ended; MCQ: multiple-choice.
% }

\begin{table*}
    \centering
    \caption{Accuracy (\%) under $L_\infty$ and $L_2$ PGD attacks ($\epsilon_\infty=4/255$; $\epsilon_2=\epsilon_\infty\sqrt{D}$, matched maximal energy, $D$ per-model) for different models across different benchmarks. Clean accuracy is repeated in both since it's norm-independent. Models that suffer higher relation degradation ($\rho>1$, $G>0$) under any attack are highlighted with `*'}
    \label{tab:main_results}
    \small
    \setlength{\tabcolsep}{4.5pt}
    \renewcommand{\arraystretch}{1.05}
    \begin{tabular}{@{}lllccc@{\hspace{15pt}}ccc@{}}
        \toprule
        Dataset & Model & Image Type & \multicolumn{3}{c@{\hspace{10pt}}}{$L_\infty$}
        & \multicolumn{3}{c}{$L_2$} \\
        \cmidrule(lr){4-6} \cmidrule(l){7-9}

        % ==================== GQA ====================
        \multirow{11}{*}{GQA}
        & & & \textit{Object} & \textit{Relation} & \textit{Total}
        & \textit{Object} & \textit{Relation} & \textit{Total} \\
        & LLaVA-1.5
        & Clean & 85.73 & 57.33 & 63.05
        & 85.73 & 57.33 & 63.05 \\
        & * LLaVA-1.5
        & Perturbed & 70.02 & 35.90 & 42.78
        & 79.09 & 47.47 & 53.85 \\
        & SLADE
        & Clean & 85.63 & 57.39 & 63.08
        & 85.63 & 57.39 & 63.08 \\
        & * SLADE
        & Perturbed & 76.55 & 48.15 & 53.87
        & 74.72 & 44.92 & 50.93 \\
        & TeCoA
        & Clean & 74.55 & 46.46 & 52.13
        & 74.55 & 46.46 & 52.13 \\
        & * TeCoA
        & Perturbed & 73.43 & 45.00 & 50.73
        & 72.77 & 44.19 & 49.95 \\
        & FARE
        & Clean & 78.77 & 50.21 & 55.97
        & 78.77 & 50.21 & 55.97 \\
        & * FARE
        & Perturbed & 76.92 & 47.82 & 53.69
        & 75.58 & 45.87 & 51.86 \\
        & Gemma-4
        & Clean & 68.10 & 53.26 & 56.25
        & 68.10 & 53.26 & 56.25 \\
        & * Gemma-4
        & Perturbed & 59.96 & 37.74 & 42.22
        & 64.26 & 44.70 & 48.64 \\

        \addlinespace[2pt]
        \midrule
        \addlinespace[2pt]

        % ==================== Spatial-MM ====================
        \multirow{11}{*}{Spatial-MM}
        & & & \textit{Single Obj.} & \textit{Dual Obj.} & \textit{Total}
        & \textit{Single Obj.} & \textit{Dual Obj.} & \textit{Total} \\
        & LLaVA-1.5
        & Clean & 48.80 & 42.13 & 44.28
        & 48.80 & 42.13 & 44.28 \\
        & * LLaVA-1.5
        & Perturbed & 43.43 & 40.61 & 41.52
        & 48.01 & 41.18 & 43.38 \\
        & SLADE
        & Clean & 50.80 & 42.79 & 45.37
        & 50.80 & 42.79 & 45.37 \\
        & SLADE
        & Perturbed & 47.01 & 41.65 & 43.38
        & 46.22 & 41.84 & 43.25 \\
        & TeCoA
        & Clean & 44.42 & 41.08 & 42.16
        & 44.42 & 41.08 & 42.16 \\
        & * TeCoA
        & Perturbed & 45.42 & 41.27 & 42.61
        & 43.43 & 40.23 & 41.26 \\
        & FARE
        & Clean & 47.01 & 42.03 & 43.64
        & 47.01 & 42.03 & 43.64 \\
        & FARE
        & Perturbed & 44.82 & 42.79 & 43.44
        & 44.82 & 42.13 & 42.99 \\
        & Gemma-4
        & Clean & 90.24 & 68.22 & 75.32
        & 90.24 & 68.22 & 75.32 \\
        & * Gemma-4
        & Perturbed & 77.09 & 53.42 & 61.05
        & 85.06 & 59.58 & 67.80 \\

        \addlinespace[2pt]
        \midrule
        \addlinespace[2pt]

        % ==================== ReXVQA ====================
        \multirow{16}{*}{ReXVQA}
        & & & \textit{Object} & \textit{Relation} & \textit{Total}
        & \textit{Object} & \textit{Relation} & \textit{Total} \\
        % & LLaVA-1.5
        % & \textit{No image} & 28.63 & 46.42 & 29.82
        % & 28.63 & 46.42 & 29.82 \\
        & LLaVA-1.5
        & Clean & 39.47 & 44.25 & 38.92
        & 39.47 & 44.25 & 38.92 \\
        & LLaVA-1.5
        & Perturbed & 32.41 & 43.62 & 33.27
        & 26.72 & 41.49 & 28.37 \\
        & SLADE
        & Clean & 39.03 & 44.17 & 38.47
        & 39.03 & 44.17 & 38.47 \\
        & * SLADE
        & Perturbed & 35.14 & 35.81 & 34.88
        & 28.74 & 34.56 & 29.27 \\
        & TeCoA
        & Clean & 38.86 & 38.88 & 38.48
        & 38.86 & 38.88 & 38.48 \\
        & * TeCoA
        & Perturbed & 37.82 & 37.75 & 37.59
        & 36.62 & 33.90 & 35.78 \\
        & FARE
        & Clean & 37.08 & 35.07 & 36.49
        & 37.08 & 35.07 & 36.49 \\
        & FARE
        & Perturbed & 34.41 & 34.76 & 34.12
        & 26.59 & 32.35 & 26.87 \\
        % & Gemma-4
        % & \textit{No image} & 5.15 & 15.09 & 7.82
        % & 5.15 & 15.09 & 7.82 \\
        & Gemma-4
        & Clean & 77.64 & 70.02 & 76.42
        & 77.64 & 70.02 & 76.42 \\
        & Gemma-4
        & Perturbed & 17.49 & 17.77 & 17.40
        & 55.87 & 56.84 & 54.76 \\
        % & MedGemma-1
        % & \textit{No image} & 81.19 & 75.12 & 78.78
        % & 81.19 & 75.12 & 78.78 \\
        & MedGemma-1
        & Clean & 87.61 & 78.69 & 85.92
        & 87.61 & 78.69 & 85.92 \\
        & * MedGemma-1
        & Perturbed & 76.69 & 64.62 & 74.13
        & 80.02 & 68.62 & 72.28 \\

        \addlinespace[2pt]
        \midrule
        \addlinespace[2pt]

        % ==================== MedXpertQA ====================
        \multirow{16}{*}{MedXpertQA}
        & & & \textit{Understanding} & \textit{Reasoning} & \textit{Total}
        & \textit{Understanding} & \textit{Reasoning} & \textit{Total} \\
        % & LLaVA-1.5
        % & \textit{No image} & 21.30 & 18.26 & 19.10
        % & 21.30 & 18.26 & 19.10 \\
        & LLaVA-1.5
        & Clean & 23.65 & 19.99 & 21.00
        & 23.65 & 19.99 & 21.00 \\
        & LLaVA-1.5
        & Perturbed & 21.84 & 20.19 & 20.65
        & 23.83 & 20.61 & 21.50 \\
        & SLADE
        & Clean & 25.09 & 19.64 & 21.15
        & 25.09 & 19.64 & 21.15 \\
        & SLADE
        & Perturbed & 22.20 & 17.36 & 18.70
        & 17.87 & 14.04 & 15.10 \\
        & TeCoA
        & Clean & 19.49 & 17.36 & 17.95
        & 19.49 & 17.36 & 17.95 \\
        & * TeCoA
        & Perturbed & 19.13 & 16.04 & 16.90
        & 15.34 & 16.80 & 16.40 \\
        & FARE
        & Clean & 20.22 & 16.74 & 17.70
        & 20.22 & 16.74 & 17.70 \\
        & * FARE
        & Perturbed & 20.76 & 16.32 & 17.55
        & 15.34 & 10.51 & 11.85 \\
        % & Gemma-4
        % & \textit{No image} & 25.63 & 32.99 & 30.95
        % & 25.63 & 32.99 & 30.95 \\
        & Gemma-4
        & Clean & 39.17 & 37.21 & 37.75
        & 39.17 & 37.21 & 37.75 \\
        & Gemma-4
        & Perturbed & 27.26 & 31.88 & 30.60
        & 31.95 & 32.30 & 32.20 \\
        % & MedGemma-1
        % & \textit{No image} & 29.60 & 24.20 & 25.70
        % & 29.60 & 24.20 & 25.70 \\
        & MedGemma-1
        & Clean & 28.70 & 26.21 & 26.90
        & 28.70 & 26.21 & 26.90 \\
        & * MedGemma-1
        & Perturbed & 27.44 & 24.69 & 25.45
        & 28.52 & 24.55 & 25.65 \\

        \bottomrule
    \end{tabular}\vspace{-5mm}
\end{table*}

\subsection{Evaluation}

For each dataset in Table \ref{tab:datasets}, we report accuracy on the full evaluation set and separately for each within-dataset capability split: object vs. relational questions (GQA), single vs. dual-object questions (Spatial-MM), reasoning vs. understanding questions (MedXpertQA), and object detection vs. spatial reasoning questions (ReXVQA). For all combinations of model-dataset-perturbation, we experiment with both $L_\infty$ and $L_2$ norms with $\epsilon_\infty=4/255$, $\epsilon_2 = \epsilon_\infty \sqrt{D}$ where $D$ is the dimensionality of the attack space, $\alpha=\epsilon/4$ and $T=10$ steps.

Since not all datasets provide explicit object- and relation-based tasks like GQA, we consider each dataset's native task splits as a proxy for object recognition and relational understanding tasks based on the type and visual reasoning required for a consistent capability-level assessment. Specifically, we consider Spatial-MM's single-object questions as an object-level proxy, while dual-object questions that require reasoning on relationships between different entities are treated as a relation-level proxy. 
% Similarly, we treat coarse vs. granular questions in DORI as object vs relation questions.
MedXpertQA does not explicitly distinguish this way and categorizes them as understanding (primarily assessing medical knowledge) and reasoning (requiring more complex reasoning). We therefore use them as proxies for lower- and higher-order capabilities, respectively. For ReXVQA, out of five classes, we map presence assessment and negation assessment to object detection, and geometric information and location assessment to spatial reasoning. We exclude 26 questions from the ReXVQA public test split that exceed the 4,096-token context limit of LLaVA-1.5 due to seven or more associated images; this affects less than 0.075\% of questions and has negligible impact on results.

\section{Results and Discussion}

\textbf{Overall result:}
We evaluate whether adversarial image perturbations disproportionately degrade relational understanding relative to object recognition across models and tasks. Throughout our experiments, we primarily report accuracy under two conditions: clean images and adversarially perturbed images. Table \ref{tab:main_results} shows a consistent pattern for the models across all benchmarks. While both capabilities degrade under attack, the drop is generally larger for relation-based questions than for object-based ones. 

The GQA benchmark, with its native object-relation task separation, offers the clearest picture. Under the adversarial attack, all models degrade more severely on relation-based questions than on object-based ones. 

For vanilla LLaVA-1.5 under $L_\infty$ PGD attack, accuracy drops from 85.73/57.33\% (object/relation) to 70.02/35.90\%, yielding capability disparity ratio $\rho=1.36 > 1$, and capability disparity gap $G=5.72 > 0$. We observe similar degradation in the other models, as reported in Table~\ref{tab:main_results}, where an asterisk highlights cases in which the model loses more accuracy on relation-based questions than on object-based ones under attack. This pattern extends to other benchmarks, though the magnitude of degradation varies by dataset and model. For instance, on Spatial-MM, both LLaVA-1.5 and Gemma-4 show a higher degradation for dual-object questions than single-object ones, suggesting that relational vulnerability intensifies as questions require reasoning over multiple entities. 

% For a direct real-world implication on clinical diagnosis, we evaluate the domain-specific MedGemma-1 on clinical diagnosis benchmarks like ReXVQA and MedXpertQA, and observe a similar scenario. In both benchmarks, MedGemma-1 performs worse on relational and higher resasoning tasks. 

Medical benchmarks, where performance is more critical for safety-critical deployment, tell a similar story. On ReXVQA, MedGemma-1's accuracy drops from 87.61/78.69\% to 76.69/64.62\% under the $L_\infty$ attack, with greater degradation in spatial understanding than object detection, a pattern that holds under the $L_2$ attack as well.

% MedXpertQA present a different evaluates understanding and reasoning tasks, and here we see a relatively smaller absolute degradation compared to the other ones. 
MedXpertQA presents a different regime. Each question provides a rich textual clinical context (e.g., possible cause of disease, additional symptoms, known side-effects), which we find dominates model performance.
To quantify this, we evaluate the models in a text-only setting without providing any image to the models, establishing a performance floor. Models perform near this floor even on clean images, indicating limited visual contribution.
% along with our standard clean and adversarial setting for same questions. T
% This provides a floor below which the visual contribution in performance is relatively low. We find that models do not noticeably perform above the text-based floor, suggesting that textual context here contains a high contribution in performance.
Nevertheless, under the adversarial attacks, MedGemma-1 still performs worse on higher reasoning tasks compared to understanding tasks, which is consistent with our broader findings that relational and higher-order capabilities are disproportionately affected.  
% \subsection{Vulnerability Beyond Standard Object-Relation Boundary}
% \subsection{Vulnerability Across Attack Norms}
% \subsection{Controlled Capability-Level Assessment} 
% Models fine-tuned in adversarial settings are expected to perform better than the vanilla models, yet we find degradation consistent with their vanilla counterpart. 
% From Table \ref{tab:main_results}, we see in majority of the cases, adversarially fine-tuned defense models (SLADE, TeCoA, FARE) do not show noticeable improvement over their vanilla counterpart in closing the performance gap on relational and higher reasoning tasks. 

\textbf{Do defense models close the gap?}
Even though the existing defense models are optimized against adversarial perturbations, none of them properly mitigate the relational vulnerability. Our results in Table \ref{tab:main_results} show that SLADE, TeCoA, and FARE perform generally better than the vanilla LLaVA-1.5 under attack, but larger degradation in relation-based questions is still evident for most of them. 
This finding leads to a more direct analysis of whether these defense models actually reduce the object-relation gap. To measure that, we calculate the normalized residual gap on GQA, following Equations \ref{eqn:ng_obj} -- \ref{eqn:normalized_gap}. As shown in Table \ref{tab:normalized_gap}, for all three defense models, the normalized gap on relation-based questions exceeds the gap in object-based questions.
While all three defense models show a higher accuracy than LLaVA-1.5 under attack, they do not close the gap in relational degradation.
\begin{table}[t]
    \centering
    \caption{Evaluating defense models' accuracy (\%) against image perturbation with the PGD attack ($L_\infty$; $\epsilon_\infty=4/255$), reporting the normalized gap on GQA. LLaVA scores are on clean images.}
    \label{tab:normalized_gap}
   \small
    \setlength{\tabcolsep}{2.5pt}
    \renewcommand{\arraystretch}{1.25}
    
    \begin{tabular}{@{}lccc@{\hspace{4pt}}cccc@{}}
        \toprule
        Model & \multicolumn{3}{c@{\hspace{4pt}}}{Accuracy} 
        & \multicolumn{4}{c}{Normalized Gap ($\mathrm{NG}$)} \\
        \cmidrule(lr){2-4} \cmidrule(lr){5-8}
        
        & \textit{Obj.} & \textit{Rel.} & \textit{Total}
        & $\mathrm{NG}_{\mathrm{obj}}$
        & $\mathrm{NG}_{\mathrm{rel}}$
        & $\mathrm{NG}_{\mathrm{total}}$
        & $\mathrm{G}_{\mathrm{def}}$ \\
        \midrule
        LLaVA-1.5
        & 85.73 & 57.33 & 63.05
        & -- & -- & -- & -- \\
        SLADE
        & 76.55 & 48.15 & 53.87
        & 10.71 & 16.01 & 14.56 & 5.30 \\
        TeCoA
        & 73.43 & 45.00 & 50.73
        & 14.35 & 21.51 & 19.54 & 7.16 \\
        FARE
        & 76.92 & 47.82 & 53.69
        & 10.28 & 16.59 & 14.85 & 6.31 \\
        \bottomrule
    \end{tabular}\vspace{-5mm}
\end{table}
\begin{table*}[htb]
    \centering
    \caption{Accuracy (\%) degradation due to PGD attack ($L_\infty$; $\epsilon_\infty=4/255$) compared against Gaussian noise on GQA.}
    \label{tab:gaussian_noise}
    \small
    \setlength{\tabcolsep}{4.5pt}
    \renewcommand{\arraystretch}{1.25}
    \begin{tabular}{@{}llcccccccc@{}}
        \toprule
        Model & Perturbation 
        & $\operatorname{ObjAcc}_{\mathrm{clean}}$ & $\operatorname{ObjAcc}_{\mathrm{adv}}$ 
        & $\operatorname{RelAcc}_{\mathrm{clean}}$ & $\operatorname{RelAcc}_{\mathrm{adv}}$ 
        & $\Delta\operatorname{RelAcc}$ & $\Delta\operatorname{ObjAcc}$ & $\rho$ & $G$ \\
        \midrule

        % ==================== LLaVA-1.5 ====================
        \multirow{2}{*}{LLaVA-1.5}
        & PGD   & 85.73 & 70.02 & 57.33 & 35.90 & 21.43 & 15.71 & 1.36  & 5.72 \\
        & Noise & 85.73 & 85.77 & 57.33 & 57.28 & 0.05  & -0.04 & -1.25 & 0.09 \\
        \midrule
        
        \addlinespace[2pt]
        
        % ==================== SLADE ====================
        \multirow{2}{*}{SLADE}
        & PGD   & 85.63 & 76.55 & 57.39 & 48.15 & 9.24  & 9.08  & 1.02  & 0.16 \\
        & Noise & 85.63 & 85.47 & 57.39 & 57.36 & 0.03  & 0.16  & 0.19  & -0.13 \\
        \midrule
        
        \addlinespace[2pt]
        
        % ==================== TeCoA ====================
        \multirow{2}{*}{TeCoA}
        & PGD   & 74.55 & 73.43 & 46.46 & 45.00 & 1.46  & 1.12  & 1.30  & 0.34 \\
        & Noise & 74.55 & 74.63 & 46.46 & 46.45 & 0.01  & -0.08 & -0.13 & 0.09 \\
        \midrule
        
        \addlinespace[2pt]
        
        % ==================== FARE ====================
        \multirow{2}{*}{FARE}
        & PGD   & 78.77 & 76.92 & 50.21 & 47.82 & 2.39  & 1.85  & 1.29  & 0.54 \\
        & Noise & 78.77 & 78.69 & 50.21 & 50.24 & -0.03 & 0.08  & -0.38 & -0.11 \\
        \midrule

        \addlinespace[2pt]

        % ==================== Gemma-4 ====================
        \multirow{2}{*}{Gemma-4}
        & PGD   & 68.10 & 59.96 & 53.26 & 37.74 & 15.52 & 8.14  & 1.91  & 7.38 \\
        & Noise & 68.10 & 68.16 & 53.26 & 53.02 & 0.24  & -0.06 & -4.00 & 0.30 \\

        \bottomrule
    \end{tabular}
\end{table*} 
\begin{table*}[htb]
    \centering
    \caption{Accuracy(\%), accuracy gap (percentage points), and cluster permutation test of LLaVA-1.5 on the controlled GQA subset under $L_\infty$ and $L_2$ PGD attacks ($\epsilon_\infty=4/255$; $\epsilon_2=\epsilon_\infty\sqrt{D}$).}
    \label{tab:controlled_subset}
    \small
    \setlength{\tabcolsep}{4.5pt}
    \renewcommand{\arraystretch}{1.25}
    \begin{tabular}{@{}llcccccccc@{}}
        \toprule
        Image Type & PGD Norm 
        & Object & Relation & Total 
        & $\Delta\operatorname{ObjAcc}$ & $\Delta\operatorname{RelAcc}$ & $G$ 
        & 95\% CI & $p$ \\
        \midrule
        
        Clean & -- 
        & 85.45 & 56.77 & 71.11 
        & -- & -- & -- & -- & -- \\
        
        \addlinespace[4pt]
        
        \multirow{2}{*}{Adversarial}
        & $L_\infty$ 
        & 71.74 & 37.51 & 54.63 
        & 13.71 & 19.26 & 5.55 
        & $[1.25, 9.68]$ & 0.0095 \\
        & $L_2$ 
        & 80.24 & 47.69 & 63.96 
        & 5.21  & 9.08  & 3.87 
        & $[0.67, 7.08]$ & 0.0192 \\
        
        \bottomrule
    \end{tabular}\vspace{-3mm}
\end{table*}

\textbf{Is the vulnerability targeted or random?} A follow-up concern is whether this gap is a consequence of an adversarial attack or random pixel-level image corruption. To evaluate this, we provide an additional controlled experiment on GQA comparing performance under a PGD attack and perturbation with random Gaussian noise. For each model (LLaVA-1.5, SLADE, TeCoA, FARE, and Gemma-4), we measure accuracy with clean, PGD-perturbed, and noise-perturbed images separately, and measure both $\rho$ and $G$.

As reported in Table \ref{tab:gaussian_noise}, no model shows meaningful performance degradation under random noise, in contrast to the values observed under PGD. This suggests that our prior finding on the sensitivity in relational understanding is not random, but rather the consequence of attack-directed optimization, containing the potential for different forms of exploitation.

\textbf{Ruling out image complexity as a confound.} 
% Another concern is that relational question might be associated with more difficult or visually cluttered scenes. 
A potential confound is that relational questions might be associated with more visually complex or cluttered scenes, making reasoning inherently harder regardless of any attack. To rule this out, we conduct a controlled experiment with LLaVA-1.5 on a subset of GQA that contains the same number of object- and relation-level questions associated with each image ($\lvert \mathcal{X} \rvert = 637$ images, $\lvert \mathcal{Q} \rvert = 2,378$ questions), so both capabilities are evaluated under identical visual conditions.

% Another concern is that relational question might be associated with more visually complex or cluttered scenes.
% The intuition here is that, the more cluttered the scene is, or more objects a scene contains, the more difficult it can be for the models to reason over relationships between them. To rule this out, we conduct a controlled experiment with LLaVA-1.5 on a paired subset of GQA that contains same number of object- and relation-level questions.

The results in Table~\ref{tab:controlled_subset} show that LLaVA-1.5 still loses more on relation-level questions than object-level ones under both attack norms. A cluster permutation test confirms this difference is statistically significant ($p < 0.02$) under both $L_\infty$ and $L_2$ attacks, surviving Bonferroni correction.

Overall, the results give us a consistent picture. Adversarial perturbations can selectively degrade relational understanding beyond object recognition.  This gap is consistent across attack norms, domains, scene complexities, and remains statistically significant. Taken together, these findings motivate optimizing for capability-level robustness as a distinct aspect of adversarial fine-tuning.

\section{Conclusion}
We identify and characterize a capability-level vulnerability of LVLMs to adversarial perturbation. Relational understanding is generally more susceptible to perturbations than object recognition. Across general-purpose and domain-adapted LVLMs, as well as adversarially fine-tuned LVLMs, benchmarks, and attack norms, relation-level reasoning frequently shows greater performance degradation than object-level understanding. Our controlled evaluation provides further evidence of this disparity while reducing potential confounding from differences in the evaluated images, and experiments with Gaussian noise suggest that this effect is due to adversarial attack rather than arbitrary pixel-level corruption. Moreover, although adversarially fine-tuned defense models improve overall robustness, they do not eliminate the gap between object- and relation-level degradation. Together, these findings suggest that aggregate robustness metrics can obscure differences in how adversarial attacks affect individual visual capabilities. We therefore argue that evaluating and optimizing LVLM robustness at the capability level, supported by additional diagnostic metrics like capability disparity ratio and gap that we introduce, is helpful for identifying failure modes that can otherwise remain hidden under holistic evaluation and for developing defenses that provide more balanced robustness in real-world and safety-critical applications. 

\section*{Acknowledgment}
Part of the work used the Delta system at the National Center of Supercomputing Applications (NCSA), allocated under the NSF NAIRR Pilot Program (\#NAIRR260259). Generative AI tools were used for proofreading, sentence enhancement, and grammatical correction in this paper.

{
    \small
    \bibliographystyle{ieeenat_fullname}
    \bibliography{main}

@String(CVPR= {IEEE Conf. Comput. Vis. Pattern Recog.})

@String(ICLR = {Int. Conf. Learn. Represent.})

@String(AAAI = {AAAI})

@String(CVPR  = {CVPR})

@String(ICLR  = {ICLR})

@inproceedings{hudson2019gqa,
  title={{GQA}: A new dataset for real-world visual reasoning and compositional question answering},
  author={Hudson, Drew A and Manning, Christopher D},
  booktitle={2019 IEEE/CVF Conference on Computer Vision and Pattern Recognition (CVPR)},
  pages={6693--6702},
  year={2019},
  organization={IEEE}
}

@inproceedings{shiri-etal-2024-empirical,
    title = "An Empirical Analysis on Spatial Reasoning Capabilities of Large Multimodal Models",
    author = "Shiri, Fatemeh  and
      Guo, Xiao-Yu  and
      Far, Mona Golestan  and
      Yu, Xin  and
      Haf, Reza  and
      Li, Yuan-Fang",
    editor = "Al-Onaizan, Yaser  and
      Bansal, Mohit  and
      Chen, Yun-Nung",
    booktitle = "Proceedings of the 2024 Conference on Empirical Methods in Natural Language Processing",
    month = nov,
    year = "2024",
    address = "Miami, Florida, USA",
    publisher = "Association for Computational Linguistics",
    url = "https://aclanthology.org/2024.emnlp-main.1195/",
    doi = "10.18653/v1/2024.emnlp-main.1195",
    pages = "21440--21455"
}

@article{tasnim2025seeing,
  title={Seeing Isn't Orienting: A Cognitively Informed Hierarchical Benchmark for Object Orientation in MLLMs},
  author={Tasnim, Nazia and Nichols, Keanu and Yan, Yuting and Ikechukwu, Nicholas and Zou, Elva and Ghadiyaram, Deepti and Plummer, Bryan A},
  journal={arXiv preprint arXiv:2505.21649},
  year={2025}
}

@inproceedings{MedXpertQA,
author = {Zuo, Yuxin and Qu, Shang and Li, Yifei and Chen, Zhangren and Zhu, Xuekai and Hua, Ermo and Zhang, Kaiyan and Ding, Ning and Zhou, Bowen},
title = {{MedXpertQA}: Benchmarking Expert-level Medical Reasoning and Understanding},
year = {2025},
publisher = {JMLR},
booktitle = {Proceedings of the 42nd International Conference on Machine Learning},
articleno = {3257},
numpages = {30},
location = {Vancouver, Canada},
series = {ICML'25}
}

@inproceedings{pal2025rexvqa,
  title={{ReXVQA}: A large-scale visual question answering benchmark for generalist chest x-ray understanding},
  author={Pal, Ankit and Lee, Jung-Oh and Zhang, Xiaoman and Sankarasubbu, Malaikannan and Roh, Seunghyeon and Kim, Won Jung and Lee, Meesun and Rajpurkar, Pranav},
  booktitle={Biocomputing 2026: Proceedings of the Pacific Symposium},
  pages={251--264},
  year={2025},
  organization={World Scientific}
}

@INPROCEEDINGS{slade,
  author={Hossain, Md Zarif and Imteaj, Ahmed},
  booktitle={2025 IEEE/CVF Conference on Computer Vision and Pattern Recognition (CVPR)}, 
  title={{SLADE}: Shielding against Dual Exploits in Large Vision-Language Models}, 
  year={2025},
  volume={},
  number={},
  pages={24244-24254},
  doi={10.1109/CVPR52734.2025.02258}
}

@inproceedings{deng2009imagenet,
  title= {{ImageNet}: A large-scale hierarchical image database},
  author={Deng, Jia and Dong, Wei and Socher, Richard and Li, Li-Jia and Li, Kai and Fei-Fei, Li},
  booktitle={2009 IEEE Conference on Computer Vision and Pattern Recognition},
  pages={248--255},
  year={2009},
  organization={Ieee}
}

@inproceedings{liu2024improved,
  title={Improved Baselines with Visual Instruction Tuning},
  author={Liu, Haotian and Li, Chunyuan and Li, Yuheng and Lee, Yong Jae},
  booktitle={2024 IEEE/CVF Conference on Computer Vision and Pattern Recognition (CVPR)},
  pages={26286--26296},
  year={2024},
  organization={IEEE}
}

@article{liu2023visual,
  title={Visual instruction tuning},
  author={Liu, Haotian and Li, Chunyuan and Wu, Qingyang and Lee, Yong Jae},
  journal={Advances in Neural Information Processing Systems},
  volume={36},
  pages={34892--34916},
  year={2023}
}

@InProceedings{pmlr-v235-karamcheti24a,
  title = 	 {Prismatic {VLM}s: Investigating the Design Space of Visually-Conditioned Language Models},
  author =       {Karamcheti, Siddharth and Nair, Suraj and Balakrishna, Ashwin and Liang, Percy and Kollar, Thomas and Sadigh, Dorsa},
  booktitle = 	 {Proceedings of the 41st International Conference on Machine Learning},
  pages = 	 {23123--23144},
  year = 	 {2024},
  editor = 	 {Salakhutdinov, Ruslan and Kolter, Zico and Heller, Katherine and Weller, Adrian and Oliver, Nuria and Scarlett, Jonathan and Berkenkamp, Felix},
  volume = 	 {235},
  series = 	 {Proceedings of Machine Learning Research},
  month = 	 {21--27 Jul},
  publisher =    {PMLR},
}

@article{team2026gemma,
  title={Gemma 4 Technical Report},
  author={{Gemma Team}},
  journal={arXiv preprint arXiv:2607.02770},
  year={2026}
}

@inproceedings{tecoa,
  title={Understanding zero-shot adversarial robustness for large-scale models},
  author={Mao, Chengzhi and Geng, Scott and Yang, Junfeng and Wang, Xin and Vondrick, Carl},
  booktitle={Proceedings of The International Conference on Learning Representations},
  year={2023}
}

@inproceedings{schlarmann2024robust,
  title={Robust CLIP: Unsupervised adversarial fine-tuning of vision embeddings for robust large vision-language models},
  author={Schlarmann, Christian and Singh, Naman Deep and Croce, Francesco and Hein, Matthias},
  booktitle={Proceedings of The International Conference on Machine Learning},
  year={2024},
  organization={PMLR}
}

@article{sellergren2025medgemma,
  title={Medgemma Technical Report},
  author={{Google Research} and {Google DeepMind}},
  journal={arXiv preprint arXiv:2507.05201},
  year={2025}
}

@inproceedings{radford2021learning,
  title={Learning transferable visual models from natural language supervision},
  author={Radford, Alec and Kim, Jong Wook and Hallacy, Chris and Ramesh, Aditya and Goh, Gabriel and Agarwal, Sandhini and Sastry, Girish and Askell, Amanda and Mishkin, Pamela and Clark, Jack and others},
  booktitle={International Conference on Machine Learning},
  pages={8748--8763},
  year={2021},
  organization={PMLR}
}

@misc{vicuna2023,
    title = {Vicuna: An Open-Source Chatbot Impressing GPT-4 with 90\%* ChatGPT Quality},
    url = {https://lmsys.org/blog/2023-03-30-vicuna/},
    author = {Chiang, Wei-Lin and Li, Zhuohan and Lin, Zi and Sheng, Ying and Wu, Zhanghao and Zhang, Hao and Zheng, Lianmin and Zhuang, Siyuan and Zhuang, Yonghao and Gonzalez, Joseph E. and Stoica, Ion and Xing, Eric P.},
    month = {March},
    year = {2023}
}

@article{alayrac2022flamingo,
  title={Flamingo: A visual language model for few-shot learning},
  author={Alayrac, Jean-Baptiste and Donahue, Jeff and Luc, Pauline and Miech, Antoine and Barr, Iain and Hasson, Yana and Lenc, Karel and Mensch, Arthur and Millican, Katherine and Reynolds, Malcolm and others},
  journal={Advances in Neural Information Processing Systems},
  volume={35},
  pages={23716--23736},
  year={2022}
}

@article{ye2025survey,
  title={A survey of safety on large vision-language models: Attacks, defenses and evaluations},
  author={Ye, Mang and Rong, Xuankun and Huang, Wenke and Du, Bo and Yu, Nenghai and Tao, Dacheng},
  journal={arXiv preprint arXiv:2502.14881},
  year={2025}
}

@inproceedings{li2024images,
  title={Images are Achilles’ heel of alignment: Exploiting visual vulnerabilities for jailbreaking multimodal large language models},
  author={Li, Yifan and Guo, Hangyu and Zhou, Kun and Zhao, Wayne Xin and Wen, Ji-Rong},
  booktitle={European Conference on Computer Vision},
  pages={174--189},
  year={2024},
  organization={Springer}
}

@inproceedings{lee2025does,
  title={How does vision-language adaptation impact the safety of vision language models?},
  author={Lee, Seongyun and Kim, Geewook and Kim, Jiyeon and Lee, Hyunji and Chang, Hoyeon and Park, Sue and Seo, Minjoon},
  booktitle={International Conference on Learning Representations},
  volume={2025},
  pages={53643--53681},
  year={2025}
}

@article{zou2023universal,
  title={Universal and transferable adversarial attacks on aligned language models},
  author={Zou, Andy and Wang, Zifan and Carlini, Nicholas and Nasr, Milad and Kolter, J Zico and Fredrikson, Matt},
  journal={arXiv preprint arXiv:2307.15043},
  year={2023}
}

@INPROCEEDINGS{sch2023adv,
  author={Schlarmann, Christian and Hein, Matthias},
  booktitle={2023 IEEE/CVF International Conference on Computer Vision Workshops (ICCVW)}, 
  title={On the Adversarial Robustness of Multi-Modal Foundation Models}, 
  year={2023},
  volume={},
  number={},
  pages={3679-3687},
  doi={10.1109/ICCVW60793.2023.00395}
  }

@article{dong2023robust,
  title={How robust is Google's Bard to adversarial image attacks?},
  author={Dong, Yinpeng and Chen, Huanran and Chen, Jiawei and Fang, Zhengwei and Yang, Xiao and Zhang, Yichi and Tian, Yu and Su, Hang and Zhu, Jun},
  journal={arXiv preprint arXiv:2309.11751},
  year={2023}
}

@inproceedings{luo2024image,
  title={An image is worth 1000 lies: Adversarial transferability across prompts on vision-language models},
  author={Luo, Haochen and Gu, Jindong and Liu, Fengyuan and Torr, Philip},
  year={2024},
  booktitle={International Conference on Learning Representations},
}

@inproceedings{wang2024stop,
  title={Stop reasoning! when multimodal llm with chain-of-thought reasoning meets adversarial image},
  author={Wang, Zefeng and Han, Zhen and Chen, Shuo and Xue, Fan and Ding, Zifeng and Xiao, Xun and Tresp, Volker and Torr, Philip and Gu, Jindong},
  year={2024},
  booktitle={Conference on Language Modeling},
}

@inproceedings{gao2024adversarial,
  title={Adversarial Robustness for Visual Grounding of Multimodal Large Language Models},
  author={Gao, Kuofeng and Bai, Yang and Bai, Jiawang and Yang, Yong and Xia, Shu-Tao},
  booktitle={ICLR Workshop on Reliable and Responsible Foundation Models},
  year={2024}
}

@inproceedings{NEURIPS2023_c1f0b856,
 author = {Carlini, Nicholas and Nasr, Milad and Choquette-Choo, Christopher A. and Jagielski, Matthew and Gao, Irena and Koh, Pang Wei and Ippolito, Daphne and Tramer, Florian and Schmidt, Ludwig},
 booktitle = {Advances in Neural Information Processing Systems},
 pages = {61478--61500},
 publisher = {Curran Associates, Inc.},
 title = {Are aligned neural networks adversarially aligned?},
 volume = {36},
 year = {2023}
}

@inproceedings{qi2024visual,
  title={Visual adversarial examples jailbreak aligned large language models},
  author={Qi, Xiangyu and Huang, Kaixuan and Panda, Ashwinee and Henderson, Peter and Wang, Mengdi and Mittal, Prateek},
  booktitle={Proceedings of the AAAI Conference on Artificial Intelligence},
  volume={38},
  number={19},
  pages={21527--21536},
  year={2024}
}

@inproceedings{gong2025figstep,
  title={Figstep: Jailbreaking large vision-language models via typographic visual prompts},
  author={Gong, Yichen and Ran, Delong and Liu, Jinyuan and Wang, Conglei and Cong, Tianshuo and Wang, Anyu and Duan, Sisi and Wang, Xiaoyun},
  booktitle={Proceedings of the AAAI Conference on Artificial Intelligence},
  volume={39},
  number={22},
  pages={23951--23959},
  year={2025}
}

@article{qraitem2024vision,
    title={Vision-{LLM}s Can Fool Themselves with Self-Generated Typographic Attacks},
    author={Qraitem, Maan and Tasnim, Nazia and Saenko, Kate and Plummer, Bryan A},
    journal={arXiv preprint arXiv:2402.00626},
    year={2024}
    }

@inproceedings{shayegani2024jailbreak,
  title={Jailbreak in pieces: Compositional adversarial attacks on multi-modal language models},
  author={Shayegani, Erfan and Dong, Yue and Abu-Ghazaleh, Nael},
  booktitle={International Conference on Learning Representations},
  volume={2024},
  pages={30853--30885},
  year={2024}
}

@article{li2023llava,
  title={LLaVA-Med: Training a large language-and-vision assistant for biomedicine in one day},
  author={Li, Chunyuan and Wong, Cliff and Zhang, Sheng and Usuyama, Naoto and Liu, Haotian and Yang, Jianwei and Naumann, Tristan and Poon, Hoifung and Gao, Jianfeng},
  journal={Advances in neural information processing systems},
  volume={36},
  pages={28541--28564},
  year={2023}
}

@inproceedings{moor2023med,
  title={{Med-Flamingo}: A multimodal medical few-shot learner},
  author={Moor, Michael and Huang, Qian and Wu, Shirley and Yasunaga, Michihiro and Dalmia, Yash and Leskovec, Jure and Zakka, Cyril and Reis, Eduardo Pontes and Rajpurkar, Pranav},
  booktitle={Machine Learning for Health (ML4H)},
  pages={353--367},
  year={2023},
  organization={PMLR}
}

@inproceedings{PromptSmooth,
  title={PromptSmooth: Certifying robustness of medical vision-language models via prompt learning},
  author={Hussein, Noor and Shamshad, Fahad and Naseer, Muzammal and Nandakumar, Karthik},
  booktitle={International Conference on Medical Image Computing and Computer-Assisted Intervention},
  pages={698--708},
  year={2024},
  organization={Springer}
}

@article{budathoki2025adversarial,
  title={Adversarial robustness analysis of vision-language models in medical image segmentation},
  author={Budathoki, Anjila and Dhakal, Manish},
  journal={arXiv preprint arXiv:2505.02971},
  year={2025}
}

@article{madry2017towards,
  title={Towards deep learning models resistant to adversarial attacks},
  author={Madry, Aleksander and Makelov, Aleksandar and Schmidt, Ludwig and Tsipras, Dimitris and Vladu, Adrian},
  journal={arXiv preprint arXiv:1706.06083},
  year={2017}
}

@article{meddeb2025evaluating,
  title={Evaluating the diagnostic accuracy of vision language models for neuroradiological image interpretation},
  author={Meddeb, Aymen and Rangus, Ida and Pagano, Paolo and Dkhil, Insaf and Jelassi, Soumaya and Bressem, Keno and Scheel, Michael and Wattjes, Mike P and Nagi, Sonia and Pierot, Laurent and others},
  journal={NPJ Digital Medicine},
  volume={8},
  number={1},
  pages={666},
  year={2025},
  publisher={Nature Publishing Group UK London}
}

@inproceedings{wang2024adashield,
  title={{AdaShield}: Safeguarding multimodal large language models from structure-based attack via adaptive shield prompting},
  author={Wang, Yu and Liu, Xiaogeng and Li, Yu and Chen, Muhao and Xiao, Chaowei},
  booktitle={European Conference on Computer Vision},
  pages={77--94},
  year={2024},
  organization={Springer}
}

@inproceedings{xu2024cross,
  title={Cross-modality information check for detecting jailbreaking in multimodal large language models},
  author={Xu, Yue and Qi, Xiuyuan and Qin, Zhan and Wang, Wenjie},
  booktitle={Findings of the Association for Computational Linguistics: EMNLP 2024},
  pages={13715--13726},
  year={2024}
}

@InProceedings{pmlr-v162-nie22a,
  title = 	 {Diffusion Models for Adversarial Purification},
  author =       {Nie, Weili and Guo, Brandon and Huang, Yujia and Xiao, Chaowei and Vahdat, Arash and Anandkumar, Animashree},
  booktitle = 	 {Proceedings of the 39th International Conference on Machine Learning},
  pages = 	 {16805--16827},
  year = 	 {2022},
  volume = 	 {162},
  series = 	 {Proceedings of Machine Learning Research},
  month = 	 {17--23 Jul},
  publisher =    {PMLR},
}

@inproceedings{gou2024eyes,
  title={Eyes closed, safety on: Protecting multimodal {LLMs} via image-to-text transformation},
  author={Gou, Yunhao and Chen, Kai and Liu, Zhili and Hong, Lanqing and Xu, Hang and Li, Zhenguo and Yeung, Dit-Yan and Kwok, James T and Zhang, Yu},
  booktitle={European Conference on Computer Vision},
  pages={388--404},
  year={2024},
  organization={Springer}
}

@InProceedings{spatialqa,
    author    = {Xie, Yuechen and Zhang, Xiaoyan and Shan, Yicheng and Hao, Zhu and Tang, Rui and Wei, Rong and Song, Mingli and Wan, Yuanyu and Song, Jie},
    title     = {SpatiaLQA: A Benchmark for Evaluating Spatial Logical Reasoning in Vision-Language Models},
    booktitle = {Proceedings of the IEEE/CVF Conference on Computer Vision and Pattern Recognition (CVPR)},
    month     = {June},
    year      = {2026},
    pages     = {2645-2657}
}

@article{khangaonkar2026multimodal,
  title={Multimodal Language Models Cannot Spot Spatial Inconsistencies},
  author={Khangaonkar, Om and Rad, Hadi J and Pirsiavash, Hamed},
  journal={arXiv preprint arXiv:2604.00799},
  year={2026}
}

@inproceedings{thrush2022winoground,
  title={Winoground: Probing vision and language models for visio-linguistic compositionality},
  author={Thrush, Tristan and Jiang, Ryan and Bartolo, Max and Singh, Amanpreet and Williams, Adina and Kiela, Douwe and Ross, Candace},
  booktitle={2022 IEEE/CVF Conference on Computer Vision and Pattern Recognition (CVPR)},
  pages={5228--5238},
  year={2022},
  organization={IEEE}
}

@article{vsr,
  title={Visual spatial reasoning},
  author={Liu, Fangyu and Emerson, Guy and Collier, Nigel},
  journal={Transactions of the Association for Computational Linguistics},
  volume={11},
  pages={635--651},
  year={2023},
  publisher={MIT Press One Broadway, 12th Floor, Cambridge, Massachusetts 02142, USA~…}
}

@inproceedings{chen2024spatialvlm,
  title={SpatialVLN: Endowing vision-language models with spatial reasoning capabilities},
  author={Chen, Boyuan and Xu, Zhuo and Kirmani, Sean and Ichter, Brain and Sadigh, Dorsa and Guibas, Leonidas and Xia, Fei},
  booktitle={Proceedings of the IEEE/CVF Conference on Computer Vision and Pattern Recognition (CVPR)},
  pages={14455--14465},
  year={2024}
}

@inproceedings{fu2024blink,
  title={Blink: Multimodal large language models can see but not perceive},
  author={Fu, Xingyu and Hu, Yushi and Li, Bangzheng and Feng, Yu and Wang, Haoyu and Lin, Xudong and Roth, Dan and Smith, Noah A and Ma, Wei-Chiu and Krishna, Ranjay},
  booktitle={European Conference on Computer Vision},
  pages={148--166},
  year={2024},
  organization={Springer}
}

@article{wang2024picture,
  title={Is a picture worth a thousand words? Delving into spatial reasoning for vision language models},
  author={Wang, Jiayu and Ming, Yifei and Shi, Zhenmei and Vineet, Vibhav and Wang, Xin and Li, Yixuan and Joshi, Neel},
  journal={Advances in Neural Information Processing Systems},
  volume={37},
  pages={75392--75421},
  year={2024}
}

@misc{gpt4vision,
    title = {{GPT-4V(ision)} System Card},
    url = {https://openai.com/index/gpt-4v-system-card/},
    author = {{OpenAI}},
    month = {September},
    year = {2023}
}

@misc{gemini_announcement,
    title = {Introducing Gemini: Our Largest and Most Capable AI Model},
    url = {https://blog.google/innovation-and-ai/technology/ai/google-gemini-ai/},
    author = {{Google}},
    month = {December},
    year = {2023}
}

@article{raza2025industrial,
  title={Industrial applications of large language models},
  author={Raza, Mubashar and Jahangir, Zarmina and Riaz, Muhammad Bilal and Saeed, Muhammad Jasim and Sattar, Muhammad Awais},
  journal={Scientific Reports},
  volume={15},
  number={1},
  pages={13755},
  year={2025},
  publisher={Nature Publishing Group UK London}
}

@article{lopez2026can,
  title={Can {ChatGPT} forecast stock price movements? Return predictability and large language models},
  author={Lopez-Lira, Alejandro and Tang, Yuehua},
  journal={Journal of Financial Economics},
  volume={184},
  pages={104335},
  year={2026},
  publisher={Elsevier}
}

@inproceedings{blv,
author = {Gonzalez Penuela, Ricardo E and Collins, Jazmin and Bennett, Cynthia and Azenkot, Shiri},
title = {Investigating Use Cases of {AI}-Powered Scene Description Applications for Blind and Low Vision People},
year = {2024},
isbn = {9798400703300},
publisher = {Association for Computing Machinery},
address = {New York, NY, USA},
booktitle = {Proceedings of the 2024 CHI Conference on Human Factors in Computing Systems},
articleno = {901},
numpages = {21},
location = {Honolulu, HI, USA},
series = {CHI '24}
}

@inproceedings{you2024ferret,
  title={{Ferret-UI}: Grounded mobile ui understanding with multimodal {LLM}s},
  author={You, Keen and Zhang, Haotian and Schoop, Eldon and Weers, Floris and Swearngin, Amanda and Nichols, Jeffrey and Yang, Yinfei and Gan, Zhe},
  booktitle={European Conference on Computer Vision},
  pages={240--255},
  year={2024},
  organization={Springer}
}

@article{brynjolfsson2025generative,
  title={Generative {AI} at work},
  author={Brynjolfsson, Erik and Li, Danielle and Raymond, Lindsey},
  journal={The Quarterly Journal of Economics},
  volume={140},
  number={2},
  pages={889--942},
  year={2025},
  publisher={Oxford University Press}
}

@article{garcia2024artificial,
  title={Artificial Intelligence--generated draft replies to patient inbox messages},
  author={Garcia, Patricia and Ma, Stephen P and Shah, Shreya and Smith, Margaret and Jeong, Yejin and Devon-Sand, Anna and Tai-Seale, Ming and Takazawa, Kevin and Clutter, Danyelle and Vogt, Kyle and others},
  journal={JAMA Network Open},
  volume={7},
  number={3},
  pages={e243201},
  year={2024}
}

@InProceedings{pmlr-v139-ramesh21a,
  title = 	 {Zero-Shot Text-to-Image Generation},
  author =       {Ramesh, Aditya and Pavlov, Mikhail and Goh, Gabriel and Gray, Scott and Voss, Chelsea and Radford, Alec and Chen, Mark and Sutskever, Ilya},
  booktitle = 	 {Proceedings of the 38th International Conference on Machine Learning},
  pages = 	 {8821--8831},
  year = 	 {2021},
  editor = 	 {Meila, Marina and Zhang, Tong},
  volume = 	 {139},
  series = 	 {Proceedings of Machine Learning Research},
  month = 	 {18--24 Jul},
  publisher =    {PMLR},
}

@article{yang2023exploring,
  title={Exploring diverse in-context configurations for image captioning},
  author={Yang, Xu and Wu, Yongliang and Yang, Mingzhuo and Chen, Haokun and Geng, Xin},
  journal={Advances in Neural Information Processing Systems},
  volume={36},
  pages={40924--40943},
  year={2023}
}

@article{tian2024drivevlm,
  title={{DriveVLM}: The convergence of autonomous driving and large vision-language models},
  author={Tian, Xiaoyu and Gu, Junru and Li, Bailin and Liu, Yicheng and Wang, Yang and Zhao, Zhiyong and Zhan, Kun and Jia, Peng and Lang, Xianpeng and Zhao, Hang},
  journal={arXiv preprint arXiv:2402.12289},
  year={2024}
}

@INPROCEEDINGS{anatomyvlm,
  author={Gu, Difei and Gao, Yunhe and Zhou, Mu and Metaxas, Dimitris},
  booktitle={2026 IEEE/CVF Winter Conference on Applications of Computer Vision (WACV)}, 
  title={{Anatomy-VLM}: A Fine-grained Vision-Language Model for Medical Interpretation}, 
  year={2026},
  volume={},
  number={},
  pages={2838-2847},
  doi={10.1109/WACV61042.2026.00278}
}

@article{liu2025survey,
  title={A survey of attacks on large vision--language models: Resources, advances, and future trends},
  author={Liu, Daizong and Yang, Mingyu and Qu, Xiaoye and Zhou, Pan and Cheng, Yu and Hu, Wei},
  journal={IEEE Transactions on Neural Networks and Learning Systems},
  year={2025},
  publisher={IEEE}
}

@article{kathait2024assessing,
  title={Assessing laterality errors in radiology: Comparing generative artificial intelligence and natural language processing},
  author={Kathait, Anjaneya Singh and Garza-Frias, Emiliano and Sikka, Tejash and Schultz, Thomas J and Bizzo, Bernardo and Kalra, Mannudeep K and Dreyer, Keith J},
  journal={Journal of the American College of Radiology},
  volume={21},
  number={10},
  pages={1575--1582},
  year={2024},
  publisher={Elsevier}
}

@inproceedings{benz2021robustness,
  title={Robustness may be at odds with fairness: An empirical study on class-wise accuracy},
  author={Benz, Philipp and Zhang, Chaoning and Karjauv, Adil and Kweon, In So},
  booktitle={NeurIPS 2020 Workshop on pre-registration in machine learning},
  pages={325--342},
  year={2021},
  organization={PMLR}
}

@inproceedings{xu2021robust,
  title={To be robust or to be fair: Towards fairness in adversarial training},
  author={Xu, Han and Liu, Xiaorui and Li, Yaxin and Jain, Anil and Tang, Jiliang},
  booktitle={International Conference on Machine Learning},
  pages={11492--11501},
  year={2021},
  organization={PMLR}
}

@article{zhao2023evaluating,
  title={On evaluating adversarial robustness of large vision-language models},
  author={Zhao, Yunqing and Pang, Tianyu and Du, Chao and Yang, Xiao and Li, Chongxuan and Cheung, Ngai-Man Man and Lin, Min},
  journal={Advances in Neural Information Processing Systems},
  volume={36},
  pages={54111--54138},
  year={2023}
}

@inproceedings{li2023evaluating,
  title={Evaluating object hallucination in large vision-language models},
  author={Li, Yifan and Du, Yifan and Zhou, Kun and Wang, Jinpeng and Zhao, Xin and Wen, Ji-Rong},
  booktitle={Proceedings of the 2023 Conference on Empirical Methods in Natural Language Processing},
  pages={292--305},
  year={2023}
}

@inproceedings{cai2024benchlmm,
  title={{BenchLMM}: Benchmarking cross-style visual capability of large multimodal models},
  author={Cai, Rizhao and Song, Zirui and Guan, Dayan and Chen, Zhenhao and Li, Yaohang and Luo, Xing and Yi, Chenyu and Kot, Alex},
  booktitle={European Conference on Computer Vision},
  pages={340--358},
  year={2024},
  organization={Springer}
}

@article{szegedy2013intriguing,
  title={Intriguing properties of neural networks},
  author={Szegedy, Christian and Zaremba, Wojciech and Sutskever, Ilya and Bruna, Joan and Erhan, Dumitru and Goodfellow, Ian and Fergus, Rob},
  journal={arXiv preprint arXiv:1312.6199},
  year={2013}
}
}

\newpage

% Reset sectioning
\setcounter{section}{0}
\setcounter{subsection}{0}
\setcounter{figure}{0}
\setcounter{table}{0}
\setcounter{equation}{0}

\renewcommand{\thesection}{S\arabic{section}}
\renewcommand{\thesubsection}{\thesection.\arabic{subsection}}
\renewcommand{\thefigure}{S\arabic{figure}}
\renewcommand{\thetable}{S\arabic{table}}
\renewcommand{\theequation}{S\arabic{equation}}

\maketitlesupplementary

\section{Prompt Construction}

\subsection{Overview}

Each model receives the same task instruction, formatted to the individual model's native conversation template. LLaVA-based models are instruction-tuned with a fixed system preamble and an explicit ``\texttt{\small USER: <image> ... ASSISTANT:}" turn structure\footnote{See the chat template: \href{https://huggingface.co/llava-hf/llava-1.5-7b-hf}{https://huggingface.co/llava-hf/llava-1.5-7b-hf}}. Gemma-based models are prompted through their processors' chat templates with images and texts formatted as separate content blocks\footnote{See the chat template: \href{https://huggingface.co/google/gemma-4-26B-A4B-it}{https://huggingface.co/google/gemma-4-26B-A4B-it}}. We use each model family's own format and tuning convention in our experiments.

\subsection{Open-ended Questions}

Evaluating open-ended questions requires the models to generate answers concise enough for a fair comparison. Hence, we prompt the models to generate answers in a way that's as similar to the ground truth answers in GQA. Incorporating the native system prompt for LLaVA\footnote{See the instances of the \texttt{Conversation} class: \\ \href{https://github.com/haotian-liu/LLaVA/blob/main/llava/conversation.py}{https://github.com/haotian-liu/LLaVA/blob/main/llava/conversation.py}}, we use this prompt for LLaVA-1.5 and other LLaVA-based defense models (e.g., SLADE, TeCoA, FARE) for GQA: 
\begin{mdframed}
\ttfamily
\small
A chat between a curious human and an artificial intelligence assistant. The assistant gives helpful, detailed, and polite answers to the human's questions. \\
USER: $\langle \text{image} \rangle$ \\ 
$\langle \text{question} \rangle$ \\
Answer the question using a single word or a short phrase.\\
Rules:\\
- For yes/no questions, output exactly yes or no.\\
- Otherwise, output the shortest canonical object, person, or attribute name.\\
- Drop articles (a, an, the).\\
- Drop descriptive modifiers (colors, sizes, materials, counts) unless the question asks for those attributes.\\
- Output only the answer, with no sentences, explanations, or punctuation.\\
ASSISTANT:
\end{mdframed}

For Gemma-4, we exclude the LLaVA-native system prompt (``\texttt{A chat ... human's questions.}") and use the following one in the question block for GQA, with the image provided in a separate image block:
\begin{mdframed}
\ttfamily
\small
Answer the question using a single word or a short phrase.\\
Rules:\\
- For yes/no questions, output exactly yes or no.\\
- Otherwise, output the shortest canonical object, person, or attribute name.\\
- Drop articles (a, an, the).\\
- Drop descriptive modifiers (colors, sizes, materials, counts) unless the question asks for those attributes.\\
- Output only the answer, with no sentences, explanations, or punctuation.\\
Q: $\langle \text{question} \rangle$ \\
A:
\end{mdframed}

\subsection{MCQ Questions}

For MCQ-based benchmarks, we don't need an extensive rulebook for the models to follow. Instead, we provide the questions and option IDs to the models and ask them only to generate the option ID of the correct answer, and we match that against the ground truth answer's option ID. 

For Spatial-MM and ReXVQA, we use this prompt for LLaVA-based models: 
\begin{mdframed}
\ttfamily
\small
USER: $\langle \text{image} \rangle$ \\
$\langle \text{question and options} \rangle$ \\
Answer with the letter of the correct option only (e.g. A, B, C, or D). \\
ASSISTANT:
\end{mdframed}

We use this prompt for Gemma-4's question block in Spatial-MM: 
\begin{mdframed}
\ttfamily
\small
$\langle \text{question and options} \rangle$ \\
Answer with the letter of the correct option only (e.g. A, B, C, or D). 
\end{mdframed}

and include an additional one-line instruction for Gemma-4 and MedGemma-1 while evaluating ReXVQA since both models often generated elaborated answers and diagnostic reasoning instead of strictly limiting to answering the option IDs:
\begin{mdframed}
\ttfamily
\small
$\langle \text{question and options} \rangle$ \\
Answer with only the single letter of the correct option (e.g., A, B, C, or D). Do not provide any rationale or extra text. 
\end{mdframed}

Many questions in MedXpertQA included multiple images. Likewise, we use the following prompt for LLaVA-based models: 
\begin{mdframed}
\ttfamily
\small
USER: $\langle \text{images} \rangle$ \\
$\langle \text{question and options} \rangle$ \\
Answer with the letter of the correct option only (e.g., A, B, C, D, or E). \\
ASSISTANT:
\end{mdframed}

and this prompt for Gemma-4 and MedGemma-1:
\begin{mdframed}
\ttfamily
\small
$\langle \text{question and options} \rangle$ \\
Answer with only the single letter of the correct option (e.g., A, B, C, D, or E). Do not provide any rationale or extra text.
\end{mdframed}

\section{Perturbations}
\begin{figure*}[htb]
\centering
\includegraphics[width=\textwidth]{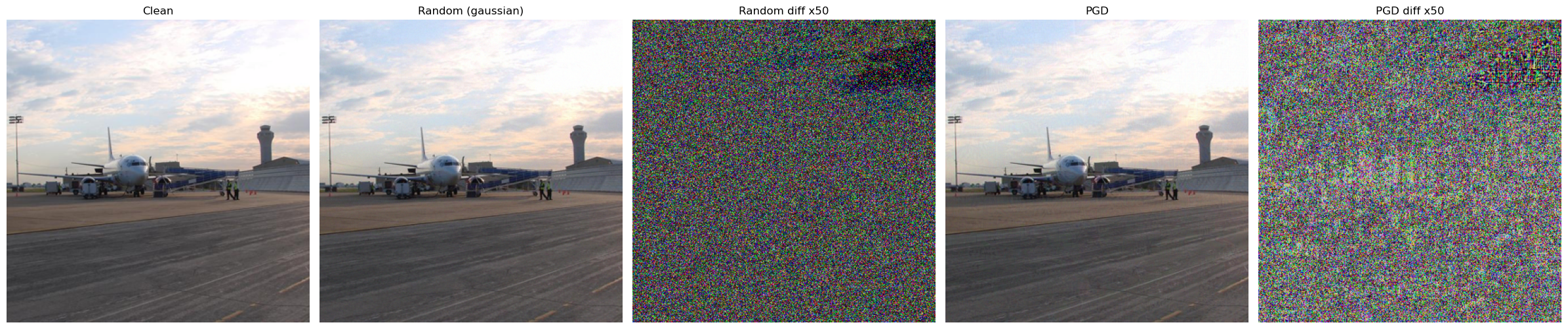}
\caption{Sample images from the GQA benchmark as clean, corrupted by Gaussian noise, and perturbed with an $L_\infty$ PGD attack. For both, we also show the absolute per-pixel residual $\lvert \text{perturbed}-\text{clean} \rvert$, amplified $50\times$ for better visibility, alongside their respective original version. Both attacks are imperceptible at their original scale.}
\label{fig:perturb-vis}
\end{figure*}

Figure \ref{fig:perturb-vis} includes visualizations of our $L_\infty$ PGD attack and Gaussian noise corruption to a sample image from the GQA benchmark. For each perturbation, we also show the absolute per-pixel residual $\lvert \text{perturbed} - \text{clean} \rvert$ with 50 times amplification for better visibility, alongside their respective original version. 

\begin{figure*}[htb]
\centering
\includegraphics[width=0.75\textwidth]{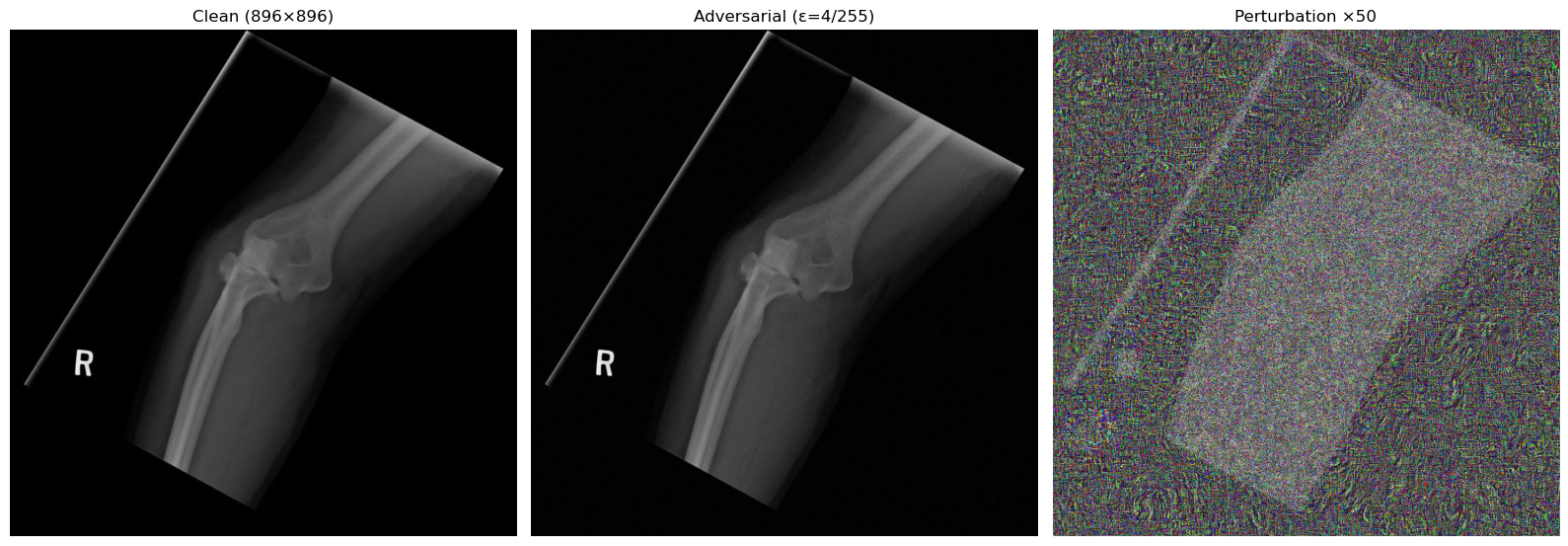}

\vspace{0.5em}
\includegraphics[width=0.75\textwidth]{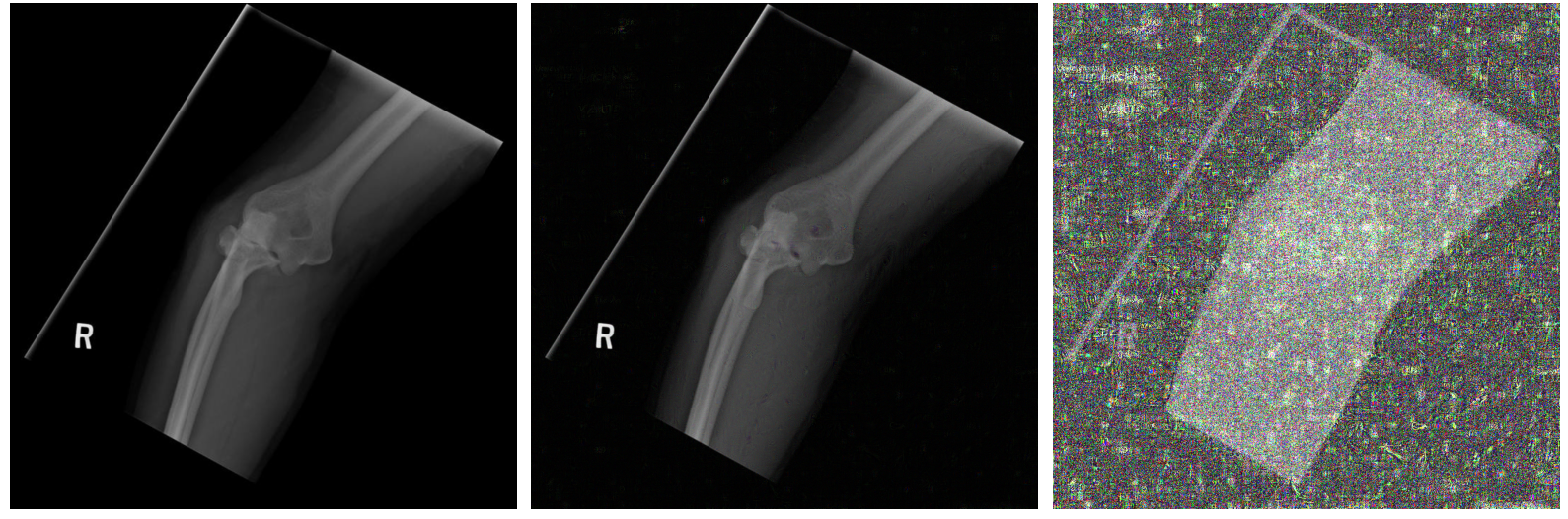}

\caption{Sample images from the MedXpertQA benchmark. Top row: clean image, the same image perturbed with an $L_\infty$ PGD attack, and the absolute per-pixel residual $\lvert \text{perturbed}-\text{clean} \rvert$, amplified $50\times$ for better visibility. Bottom row: the corresponding $L_2$ PGD attack. Both attacks are imperceptible at their original scale.}
\label{fig:perturb-vis-medx}
\end{figure*}

Figure \ref{fig:perturb-vis-medx} shows sample images from the MedXpertQA benchmark containing the clean image, the same image perturbed with $L_\infty$ and $L_2$ PGD attacks, and the absolute per-pixel residual $\lvert \text{perturbed} - \text{clean} \rvert$ with 50 times amplification for better visibility. All perturbations are imperceptible to the human eye at their original scale.

% {
%     \small
%     \bibliographystyle{ieeenat_fullname}
%     \bibliography{main}
% }

\end{document}